\documentclass[lettersize,journal]{IEEEtran}
\usepackage{amsmath,amsfonts}
\usepackage{algorithmic}
\usepackage{algorithm}
\usepackage{array}
\usepackage[caption=false,font=normalsize,labelfont=sf,textfont=sf]{subfig}
\usepackage{textcomp}
\usepackage{stfloats}
\usepackage{url}
\usepackage{verbatim}
\usepackage{graphicx}
\usepackage{cite}

\usepackage{bm}
\usepackage{amsmath}
\usepackage{tabularray}
\usepackage{array}
\usepackage{pifont}
\usepackage{bbm}
\usepackage{booktabs}
\usepackage[comma,numbers,square,sort&compress]{natbib}
\usepackage{hyperref}
\hypersetup{
	colorlinks=true,
	linkcolor=blue,
	citecolor=blue,
	urlcolor=blue
}
\usepackage{CJK}
\usepackage{enumitem}
\usepackage{amsthm}
\newtheorem{assumption}{Assumption}
\newtheorem{theorem}{Theorem}

\usepackage{multirow}
\usepackage{caption}
\usepackage{cleveref}

\begin{document}

\title{OCP-GN: A Scalable Second-order Optimizer for Stochastic Optimization}

\author{Jindi Zhong, Congyaohui Yin, Zhaorong Zhang Huanshui Zhang}

\markboth{Journal of \LaTeX\ Class Files,~Vol.~14, No.~8, August~2021}%
{Shell \MakeLowercase{\textit{et al.}}: A Sample Article Using IEEEtran.cls for IEEE Journals}

\maketitle

\begin{abstract}
This paper proposes a novel second-order optimization algorithm based on the Optimal Control Principle (OCP) \cite{ocp1}, applicable to large-scale optimization problems in neural network training. The algorithm has a computational complexity of $\mathcal{O}(d)$ and strong robustness. Extensive experiments on multiple benchmarks demonstrate the significant superiority of the proposed method.
\end{abstract}

\begin{IEEEkeywords}
Deep learning, OCP optimization.
\end{IEEEkeywords}

\section{Introduction}
In the realm of large-scale stochastic optimization for neural networks, the training procedure of neural networks generally depends on general-purpose optimizers, such as stochastic gradient descent (SGD) and adaptive moment estimation (Adam). Existing optimization algorithms fail to effectively improve the convergence speed and generalization performance of neural network training. In fact, the training of a neural network can be viewed as the evolution of a high-dimensional dynamical system in parameter space along a loss landscape. Therefore, how to systematically guide this evolutionary trajectory to converge to the optimal solution region more quickly, stably, and robustly remains a critical yet underexplored problem.
	
Meanwhile, a new optimization method of OCP provides a rigorous and powerful mathematical framework, offering a novel approach for designing optimizers suitable for neural network training.

\section{The Optimization Algorithm}
This section aims to bridge the OCP method \cite{ocp1} with large-scale stochastic optimization problems for neural networks, extending the former to the latter domain. Specifically, we propose a novel optimizer whose core idea is to inherit the update rule of OCP method, while simplifying the computation of the Hessian matrix via a Gauss-Newton-Bartlett (GNB) estimator \cite{sophia} and incorporating several technical enhancements to ensure stable updates.
\subsection{Gauss-Newton-Bartlett estimator}
Building upon the general GNB estimator, we extend its formulation and explicitly state its expression for the case of the mean squared error (MSE) loss function. We define the per-sample loss function as $\Psi(\hat{y},y)$, where $\hat{y} = \ell(x)$ denotes the model's prediction for input $x$ and $y$ is the corresponding ground-truth label. For the case of MSE loss,with the coefficient $\frac{1}{2}$ introduced to simplify gradient computation, the per-sample loss is given by:
\begin{equation}
	\Psi(\ell(x),y) = \frac{1}{2} \left( \ell(x) - y \right)^2, \label{eq20}
\end{equation}
The gradient of the per-sample loss function is:
\begin{equation}
	\nabla \Psi(\ell(x),y) = [J_x \ell(x)]^{\top} (\ell(x) - y)
\end{equation}
where $J_x \ell(x)$ is the Jacobian matrix of $\ell(x)$ w.r.t $x$.

The Hessian matrix of the per-sample loss function is:
\begin{equation}
	\nabla^2 \Psi(\ell(x),y) = [J_x \ell(x)]^{\top} J_x \ell(x) + J_{xx} \ell(x) [(\ell(x) - y)] \label{Jacobin second order}
\end{equation}
where $J_{xx} \ell(x)$ is the second-order derivatives of the multi-variate function $\ell(x)$ w.r.t $x$. In neural network research, prior work has found that the second term $J_{xx} \ell(x) [(\ell(x) - y)]$ in Equation \eqref{Jacobin second order} is generally smaller than the first term $[J_x \ell(x)]^{\top} J_x \ell(x)$ and is also computationally more challenging, so it is often simplified to
\begin{equation}
	\nabla^2 \Psi(\ell(x),y) \approx [J_x \ell(x)]^{\top} J_x \ell(x)
\end{equation}
We resample the model’s predictive distribution to obtain a synthetic label $\hat{y} \sim \mathcal{N}(\ell(x), \sigma^2)$.
Because $\Psi(\ell(x),\hat{y})$ is the negative log-probability of the probabilistic model defined by the Gaussian distribution with parameter $x$, by Bartlett's second identity, we have that,
\begin{equation}
	[J_x \ell(x)]^{\top} J_x \ell(x) = \mathop{\mathbb{E}}\limits_{\hat{y} \sim \mathcal{N}(\ell(x), \sigma^2)} [\nabla \Psi(\ell(x),\hat{y}) \nabla \Psi(\ell(x),\hat{y})^{\top}]
\end{equation}
which implies that $diag([J_x \ell(x)]^{\top} J_x \ell(x)) = \mathop{\mathbb{E}}\limits_{\hat{y} \sim \mathcal{N}(\ell(x), \sigma^2)} [\nabla \Psi(\ell(x),\hat{y}) \odot \nabla \Psi(\ell(x),\hat{y})]$. Hence, the quantity $\nabla \Psi(\ell(x),\hat{y}) \odot \nabla \Psi(\ell(x),\hat{y})$ is an unbiased estimator of the Gauss-Newton matrix for the Hessian of a one-example loss $\Psi(\ell(x),y)$.

Given a mini-batch of inputs ${\{(x_n,y_n)\}}_{n=1}^{N}$. The most natural way to build an estimator for the diagonal of the Guass-Newton matrix for the Hessian of the mini-batch loss is using
\begin{equation}
	\frac{1}{N} \sum_{n=1}^{N} \nabla \Psi(\ell_n(x),\hat{y}_n) \odot \nabla \Psi(\ell_n(x),\hat{y}_n),
\end{equation}

For the loss function $\mathcal{V}(\ell(x),\hat{y}) = \frac{1}{N} \sum_{n=1}^{N}  \left(\ell_n(x) - \hat{y}_n \right)^2$, based on Bartlett's first identity, we have:
\begin{equation}
	\begin{aligned}
		&\mathop{\mathbb{E}}\limits_{\hat{y} \sim \mathcal{N}(\ell(x), \sigma^2)}[N \cdot \nabla \mathcal{V}(\ell(x), \hat{y}) \odot \nabla \mathcal{V}(\ell(x), \hat{y})] \\
		&= \mathop{\mathbb{E}}\limits_{\hat{y} \sim \mathcal{N}(\ell(x), \sigma^2)}[\frac{1}{N} \sum_{n=1}^{N} \nabla \Psi(\ell_n(x),\hat{y}_n) \odot \nabla \Psi(\ell_n(x),\hat{y}_n)]
	\end{aligned}
\end{equation}

Consequently, when operating with mini-batch stochastic gradients in deep learning, the diagonal Hessian can be approximated (e.g., via the GNB estimator) as
\begin{equation}
	H_k = g_k \odot g_k
\end{equation}

\subsection{Closed-form update of the OCP method}
According to \cite{ocp1} - \cite{ocp-ls}, OCP method is as follows:
\begin{equation}
	\begin{aligned}
		&x_{k+1} = x_{k} - \phi_{k}(x_{k}) \\
		&\phi_{l}(x_{k}) = M \nabla f(x_{k}) + (I - M \nabla^2 f(x_{k})) \phi_{l-1}(x_{k}) \\
		&\phi_{0}(x_{k}) = M \nabla f(x_{k}), \label{ocp}
	\end{aligned}
\end{equation}
where $\phi_{l}(x_{k})$ is the update amount of the inner iteration. $M$ is the hyperparameter matrix, and $\nabla f(x_{k})$ and $\nabla^2 f(x_{k})$ are the gradient and Hessian matrix at the $k$-th iteration with respect to the optimization variables $x$.

Observing Equation (9), the dominant update term is $M \nabla f(x_{k})$, where the parameter matrix $M$ plays the role of the learning rate. We aim to rewrite the above expression in a closed-form solution while separating the parameter matrix $M$ from the update magnitude as much as possible.

Starting from $\phi_{0}$ in Equation (9), write out the closed-form expression of $\phi_{l}(x_{k})$ through iteration:
\begin{equation}
	\phi_{l}(x_{k}) = M \nabla f(x_{k}) \frac{I-(I - M \nabla^2 f(x_{k}))^{l+1}}{M \nabla^2 f(x_{k})}
	\label{ocp_update}
\end{equation}
The reason for not canceling $M$ here is to separate the learning rate as much as possible.

In the framework of large-scale stochastic optimization, we need to make the following necessary simplifications and approximations for update rule \eqref{ocp_update}. First, replace the original gradient $\nabla f(x_k)$ with the bias-corrected term of the stochastic gradient. Second, replace the computation of the full Hessian matrix with the GNB estimator, while also introducing bias correction. Therefore, update rule \eqref{ocp_update} is extended within the large-scale stochastic optimization framework as follows:
\begin{equation}
	\phi_{l}(x_{k}) = M \hat{g}_{k} \frac{I-(I - M \hat{H}_{k})^{l+1}}{M \hat{H}_{k}}
	\label{ocp_sta_update}
\end{equation}
where $\hat{g}_{k}$ and $\hat{H}_{k}$ are the bias-corrected terms defined as follows:
\begin{equation}
	\begin{aligned}
		&\hat{g}_{k} = \frac{\tilde{g}_{k}}{1 - {\beta_{1}}^{k}} \\
		&\hat{H}_{k} = \frac{\tilde{H}_{k}}{1 - {\beta_{2}}^{k}},
	\end{aligned}
\end{equation}
here, $\beta_{1}$ and $\beta_{2}$ are tunable parameters. The exponential moving average terms, $\tilde{g}_{k}$ and $\tilde{H}_{k}$, are defined as follows:
\begin{equation}
	\begin{aligned}
		&\tilde{g}_{k} = \beta_{1} \tilde{g}_{k-1} + (1-\beta_{1}) g_{k} \\
		&\tilde{H}_{k} = \beta_{2} \tilde{H}_{k-1} + (1 - \beta_{2}) H_{k},
	\end{aligned}
\end{equation}
where $g_{k}$ is the stochastic gradient at the $k$-th iteration, and $H_k$ satisfies the following condition:
\begin{equation}
	H_{k} = {g_k}^2,
\end{equation}

\subsection{Clipping Stabilization Mechanism}
This subsection proposes a scalable OCP method based on the GNB estimator (OCP-GN).
From the perspective of optimization stability, this section adopts several technical schemes for the OCP-GN algorithm, mainly involving the setting of clipping thresholds. The goal is to enable the OCP-GN algorithm to exhibit stronger stability when solving large-scale stochastic optimization problems such as neural network training.

First, for the estimation of the diagonal elements of the Hessian matrix $H$, to avoid division by zero and ensure the positive definiteness of the diagonal matrix, we set a lower bound and introduce a lower bound threshold $\mu_{H}$ for the diagonal matrix, as shown below:
\begin{equation}
	{\hat{H}^c}_{\ k} = \max(\hat{H}_k, \mu_{H})
\end{equation}

Meanwhile, to ensure that the update direction strictly follows the direction of the gradient correction term, we impose both lower and upper clipping thresholds on $\kappa_k = I - M \hat{H}_{k}$, as shown below:
\begin{equation}
	{\kappa^c}_k = \max(\mu_{1}, \min(\mu_{2}, \kappa_k))
\end{equation}

We no longer iterate the inner loop of the original OCP method for $k$ times, but instead set the number of inner loop iterations as a tunable hyperparameter.
Although the parameter matrix $M$ can be adjusted arbitrarily, in large-scale stochastic optimization problems, we introduce the learning rate $\alpha$ and describe the parameter matrix $M$ as $\alpha \cdot I$, where $I$ is the identity matrix.

To prevent overfitting and enhance the generalization ability of the model, we incorporate weight decay as a regularization technique during training.
\begin{equation}
	{x^c}_k = x_k(1 - \alpha \lambda)
\end{equation}
where $\lambda$ is the weight decay coefficient.

The OCP-GN algorithm is as follows:
\begin{equation}
	x_{k+1} = {x^c}_k + \alpha \max(-1, \min(1, \hat{g}_{k} \frac{I-({\kappa^c}_k)^{l+1}}{\alpha {\hat{H}^c}_{\ k}}))
\end{equation}

The implementation procedure of the above algorithm is illustrated in the following pseudocode:
\begin{algorithm}
	\caption{OCP-GN} \label{alg:spatial averaging}
	\begin{algorithmic}[1]
		\REQUIRE Total iterations $T$, decay factor $\beta_1$ and $\beta_2$, weight decay coefficient $\lambda$, learning rate $\alpha$, initial parameters $\theta_0$, bound threshold $\mu_H$, $\mu_1$ and $\mu_2$, the maximum number of inner loop iterations $b$.
		\ENSURE Optimized parameters $x_T$
		
		\STATE \textbf{Initialization}
		\STATE Initialize $\tilde{g}_0 \gets 0$, $k \gets 0$
		
		\WHILE{$k < T$}
		\STATE Compute stochastic gradient $g_k$ using mini-batches
		\STATE Compute $H_{k} = {g_k}^2$
		
		\STATE $\tilde{g}_{k} \gets \beta_{1} \tilde{g}_{k-1} + (1-\beta_{1}) g_{k}$
		\STATE $\hat{g}_{k} \gets \frac{\tilde{g}_{k}}{1 - \beta_{1}^{k}}$
		\STATE $\tilde{H}_{k} \gets \beta_{2} \tilde{H}_{k-1} + (1 - \beta_{2}) H_{k}$
		\STATE $\hat{H}_{k} \gets \frac{\tilde{H}_{k}}{1 - {\beta_{2}}^{k}}$
		\STATE $l \gets 0$
		\WHILE{$l < b$}
		\STATE $l \gets l+1$
		\STATE $	x_{k+1} \gets {x^c}_k + \alpha \max(-1, \min(1, \hat{g}_{k} \frac{I-({\kappa^c}_k)^{l+1}}{\alpha {\hat{H}^c}_{\ k}}))$
		\ENDWHILE
		\STATE $k \gets k+1$
		\ENDWHILE
		
		\RETURN $x_{T}$
	\end{algorithmic}
\end{algorithm}

To provide a reference for theoretical completeness, we consider a more general parent algorithm and analyze its convergence rate. To this end, we first introduce the following reasonable assumptions.
\begin{assumption} \label{ass:main}
	The following assumptions hold throughout our analysis:
	\begin{enumerate}[label=(A\arabic*), leftmargin=*, nosep]
		\item \label{ass:a1} Let the objective function $f(x): \mathbb{R}^{d+2} \to \mathbb{R}$ be continuously differentiable, and its gradient be $L$-Lipschitz continuous, i.e., there exists a constant $\beta > 0$ such that
		\begin{equation}
			f(y) \leq f(x) + \nabla f(x)^{\top} (y-x) + \frac{\beta}{2} \| y-x\|_2^2, \ \forall x,y \in \mathbb{R}^{d+2}.
		\end{equation}		
		\item \label{ass:a2} Let $f : \mathbb{R}^{d+2} \to \mathbb{R}$ be continuously differentiable and bounded below. There exists a constant $\mu > 0$ such that for all $x \in \mathbb{R}^{d+2}$,
		\[
		\frac12 \| g_k \|_2^2 \ge \mu \, ( f(x_k) - f^* ).
		\tag{PL}
		\]
		
		\item \label{ass:a3} Let $H_k$ be symmetric.
		There exists a scalar $\alpha > 0$ such that for the diagonal matrix $M = \alpha I$, 
		and for a given iteration number $k$, we have
		\[
		\big[\, (I - \alpha H_k)^{k+1} \,\big]_{ii} < 1, \qquad \forall i, \ \forall k.
		\]
		Equivalently,
		\[
		\big[\, I - (I - \alpha H_k)^{k+1} \,\big]_{ii} > 0, \qquad \forall i, \ \forall k.
		\]
		The constants $\alpha$, $\beta$, and $\mu$ appearing in this work satisfy
		\[
		\beta < 2\alpha \quad \text{and} \quad \mu \leq \frac{\alpha \beta}{2\alpha - \beta}.
		\]
	\end{enumerate}
\end{assumption}

Subsequently, a weight decay term is incorporated into the objective function, and the problem is reformulated as follows.
\begin{equation}
	\min_{x} f(x) = \min_{x}\mathcal{L}(x) + \frac{\lambda}{2} \|x\|^2,
\end{equation}

\begin{theorem} \label{thm:convergence}
	Suppose that Assumption \ref{ass:main} holds. The proposed algorithm generates a sequence $\{x_k\}$ that converges to the optimum $x^{*}$ asymptotically at a linear rate. Specifically, there exists a constant $\rho_{\infty} \in [0,1)$ such that
	\begin{equation}
		f(x_k) - f(x^*) = \mathcal{O}(\rho_{\infty}^k),
	\end{equation}
	where $\rho_{\infty}$ represents the asymptotic linear convergence rate of the algorithm.
\end{theorem}

The proof of the theorem above mainly depends on the equaivalent iterative scheme of the algorithm. It includes an adaptive step size and Newton direction. By leveraging the diagonal structure of the Hessian matrix and assuming a Lipschitz continuous gradient of the objective function, the update in each iteration is rigorously quantified. Associated with the Polyak–Łojasiewicz condition, the algorithm is proven to achieve linear convergence and an explicit expression for the convergence rate is provided.

\section{Experimetal Results}
\subsection{Experimental Environment}
All experiments were conducted in an environment utilizing PyTorch 2.5.1 (with CUDA 12.8) and Python 3.10.19.
The hardware platform featured an NVIDIA RTX A6000 GPU (48GB VRAM) and dual Intel Xeon Platinum 8380
CPU, under the Linux operating system.

\subsection{Experimental Setup}
The hyperparameters of OCP-GN were configured as
follows: total iteration T was set to 200, learning rate $\alpha$ was set to 0.0001, weight decay coefficient $\lambda$ to 0.001, bound threshold $\mu_{H}$ to $10^{-8}$, $\mu_1$ to 0.2, $\mu_2$ to 0.8, $\beta_1$ and $\beta_2$ to 0.9 and 0.999, respectively. Additionally, a cosine decay strategy was employed for the learning rate, and the same setting was applied to the baseline optimizer AdamW.

\subsection{Classification Tasks}
\begin{figure*}[htbp]
	\centering
	\subfloat[ViT, CIFAR-10]{
		\includegraphics[width=0.23\textwidth]{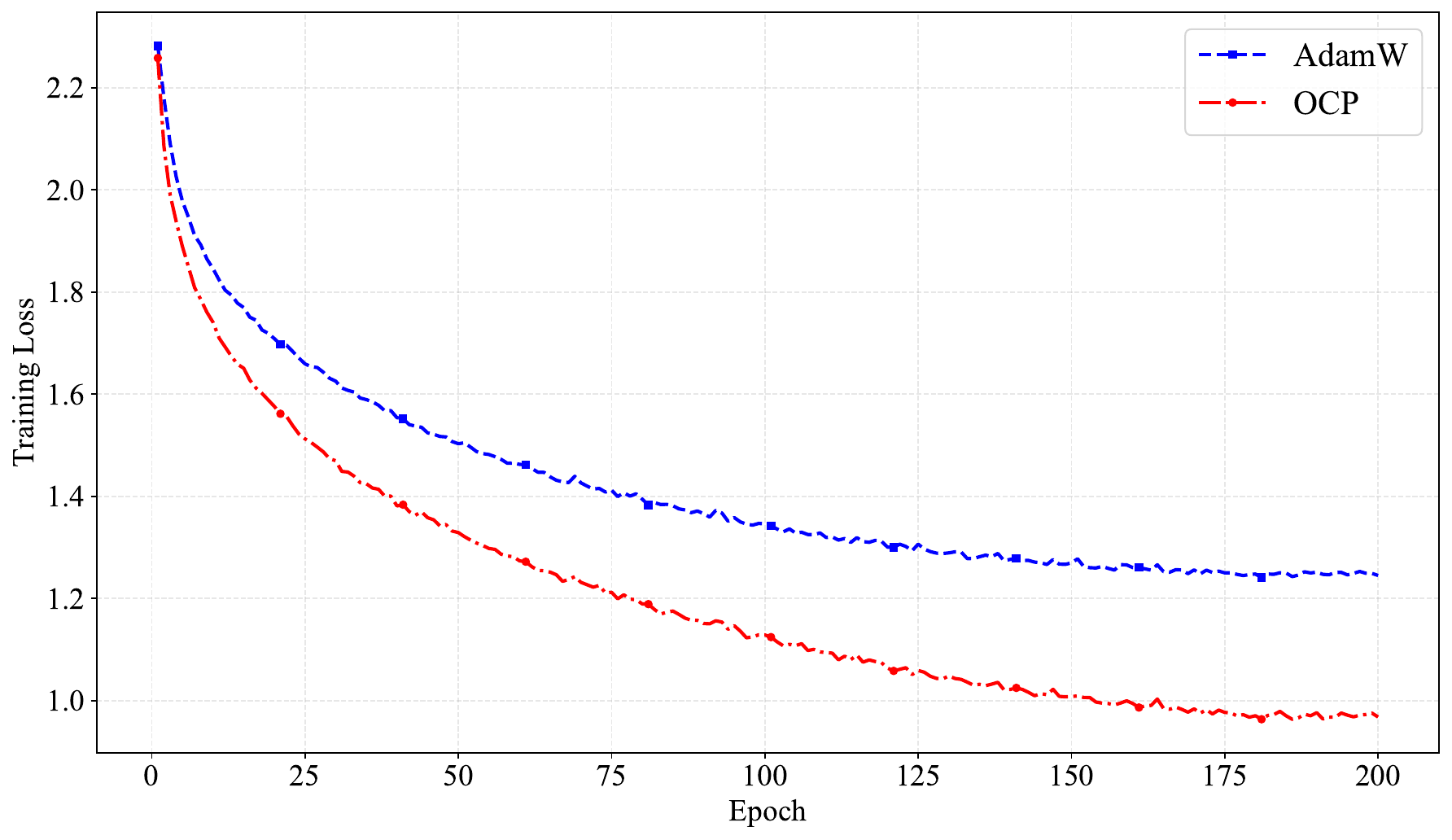}}
	\hspace{0.01\textwidth}
	\subfloat[ResNet-34, CIFAR-10]{
		\includegraphics[width=0.23\textwidth]{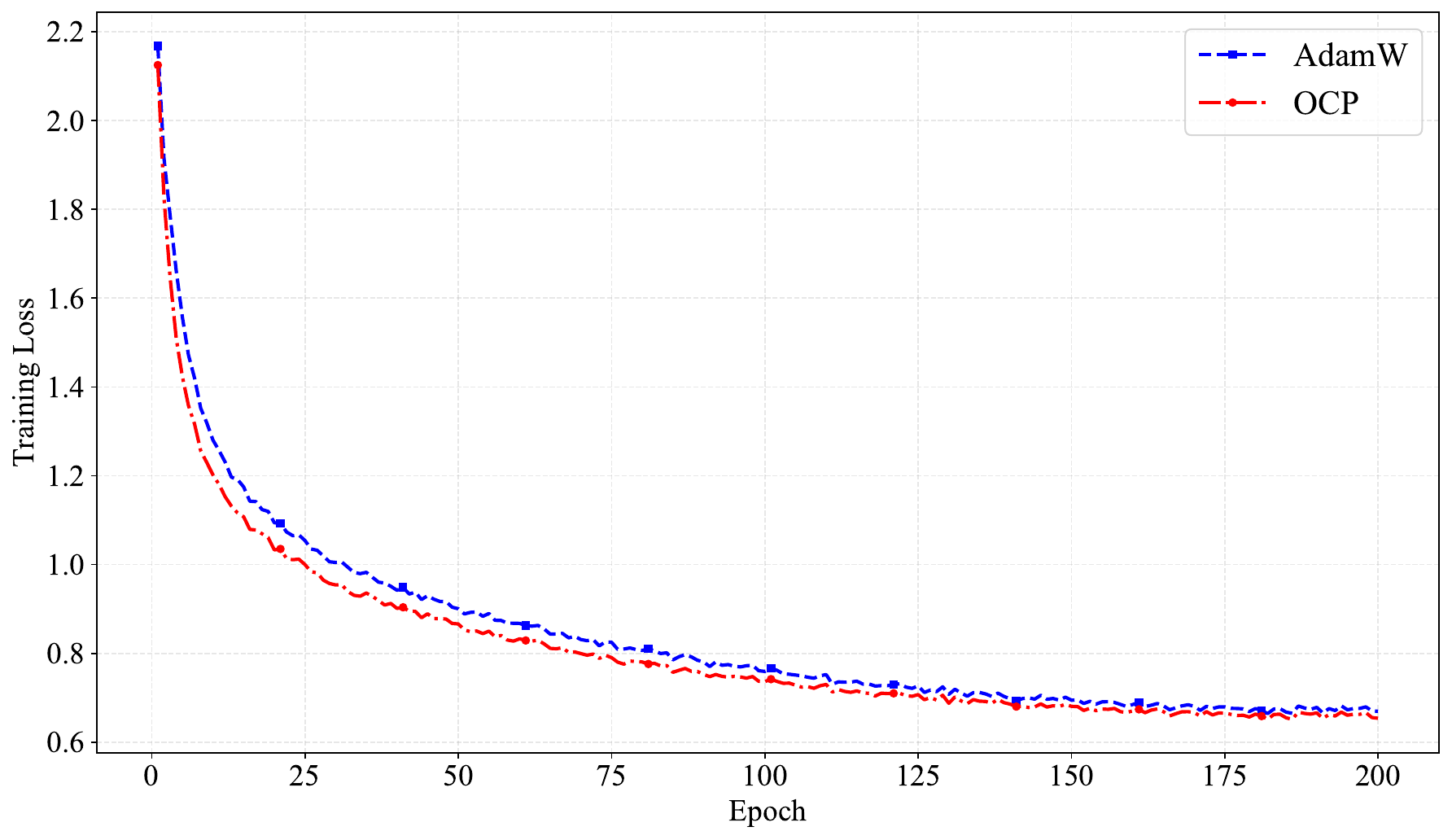}}
	\hspace{0.01\textwidth}
	\subfloat[ViT, CIFAR-100]{
		\includegraphics[width=0.23\textwidth]{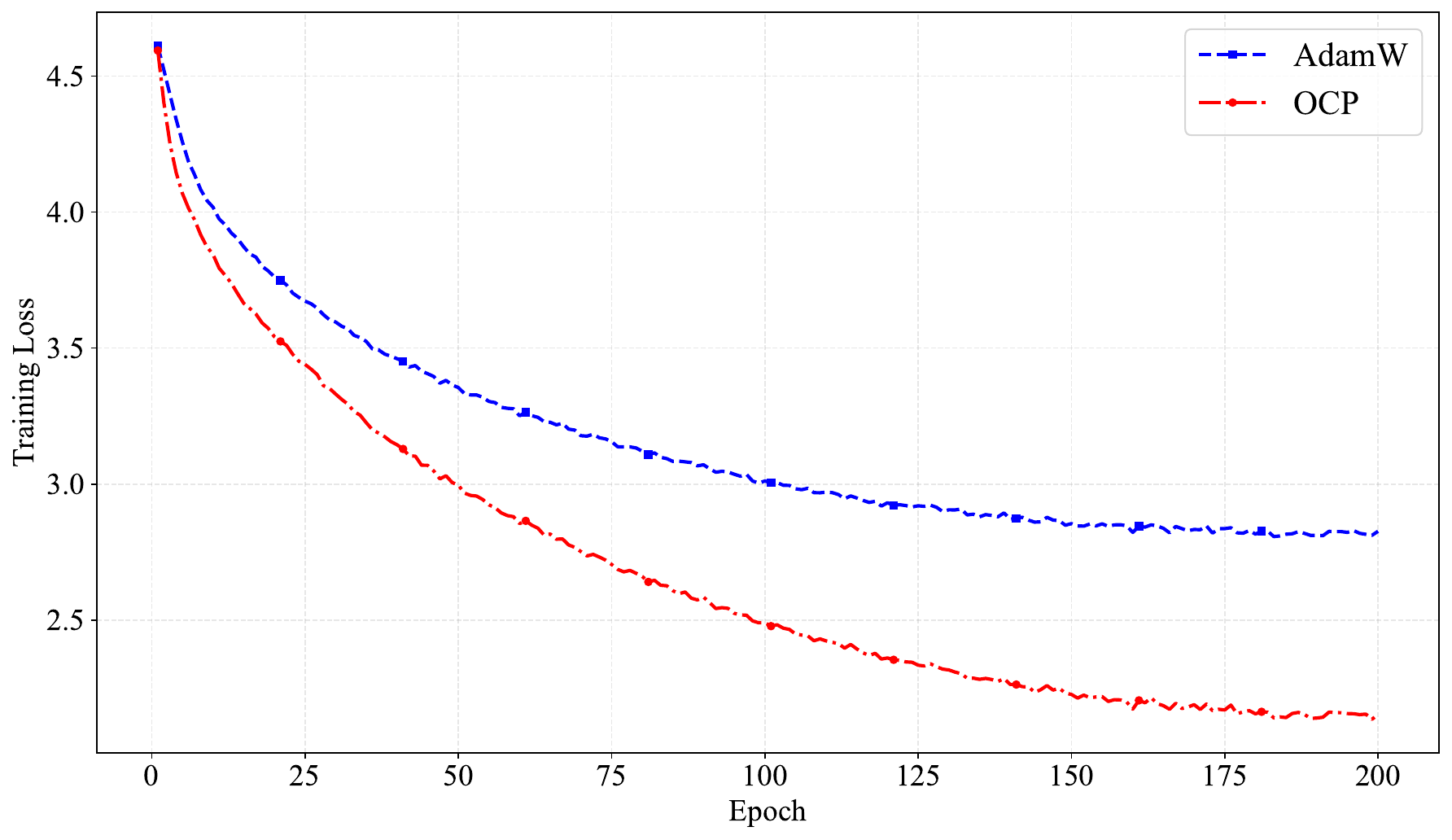}}
	\hspace{0.01\textwidth}
	\subfloat[ResNet-34, CIFAR-100]{
		\includegraphics[width=0.23\textwidth]{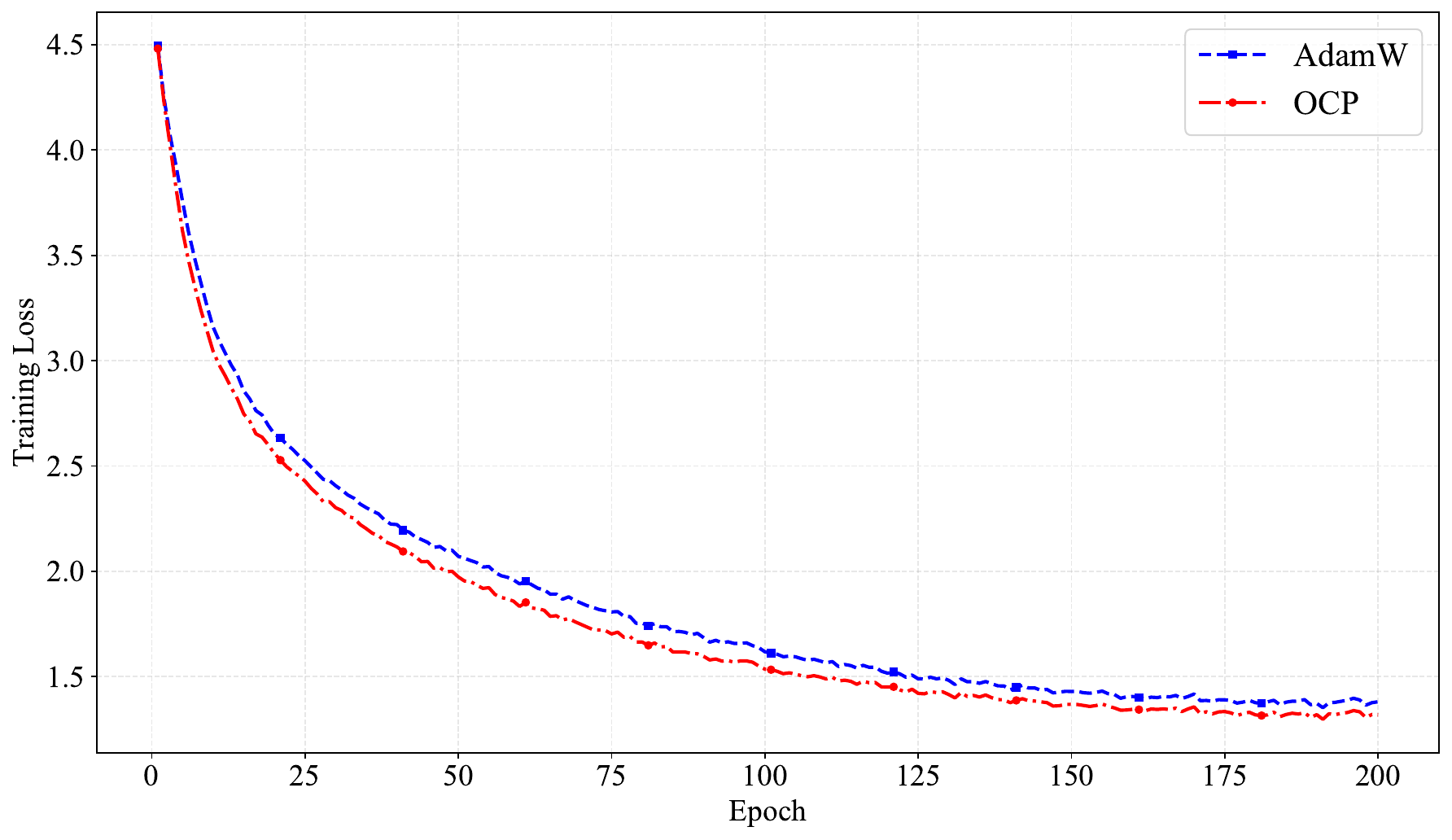}}
	
	\caption{Comparison of train loss between two algorithms.
		(a) ViT, CIFAR-10, (b) ResNet-34, CIFAR-10, (c) ViT, CIFAR-100, 
		(d) ResNet-34, CIFAR-100.}
		\label{fig:train_loss}
\end{figure*}

\begin{figure*}[htbp]
	\centering
	\subfloat[ViT, CIFAR-10]{
		\includegraphics[width=0.23\textwidth]{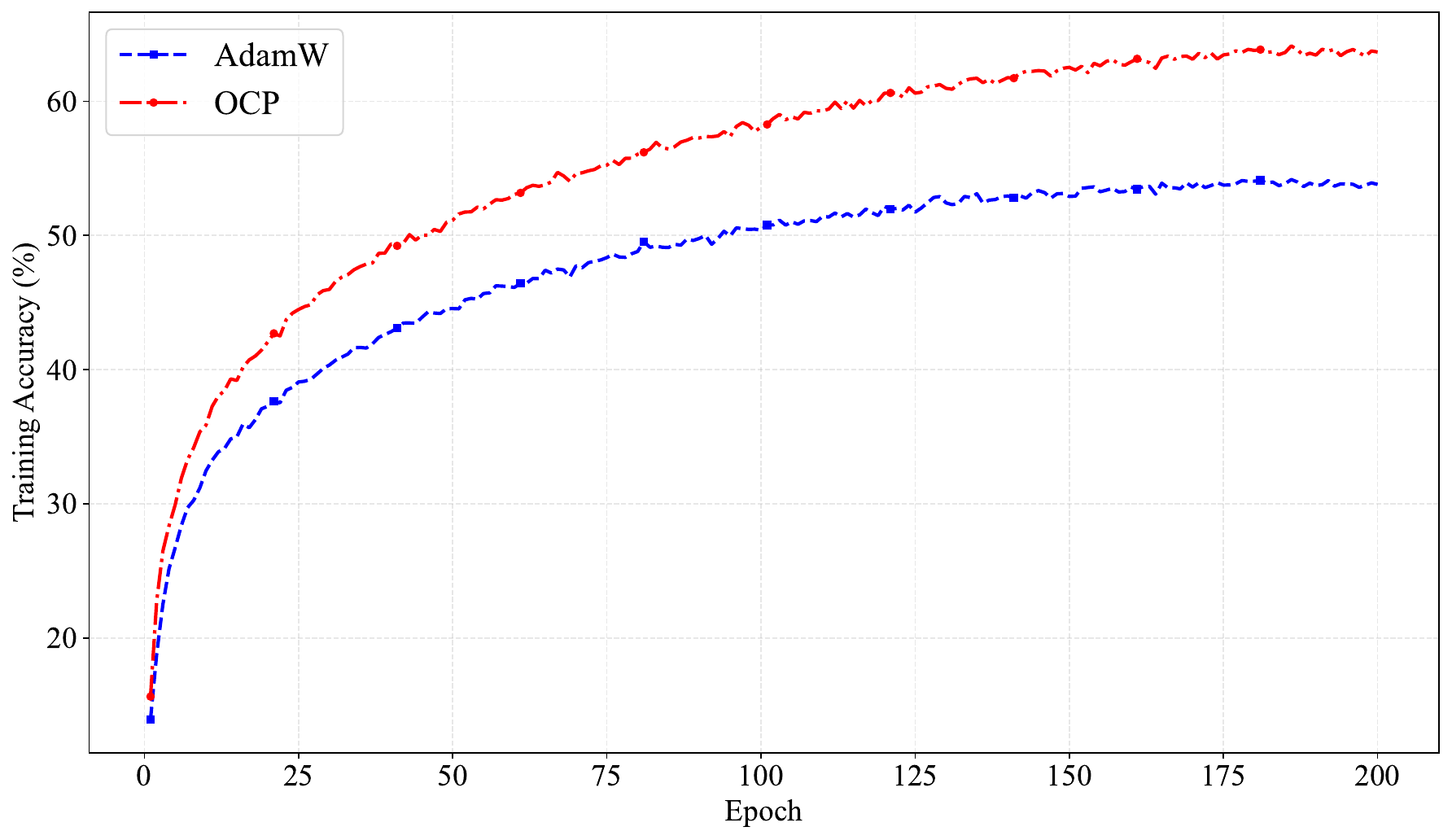}}
	\hspace{0.01\textwidth}
	\subfloat[ResNet-34, CIFAR-10]{
		\includegraphics[width=0.23\textwidth]{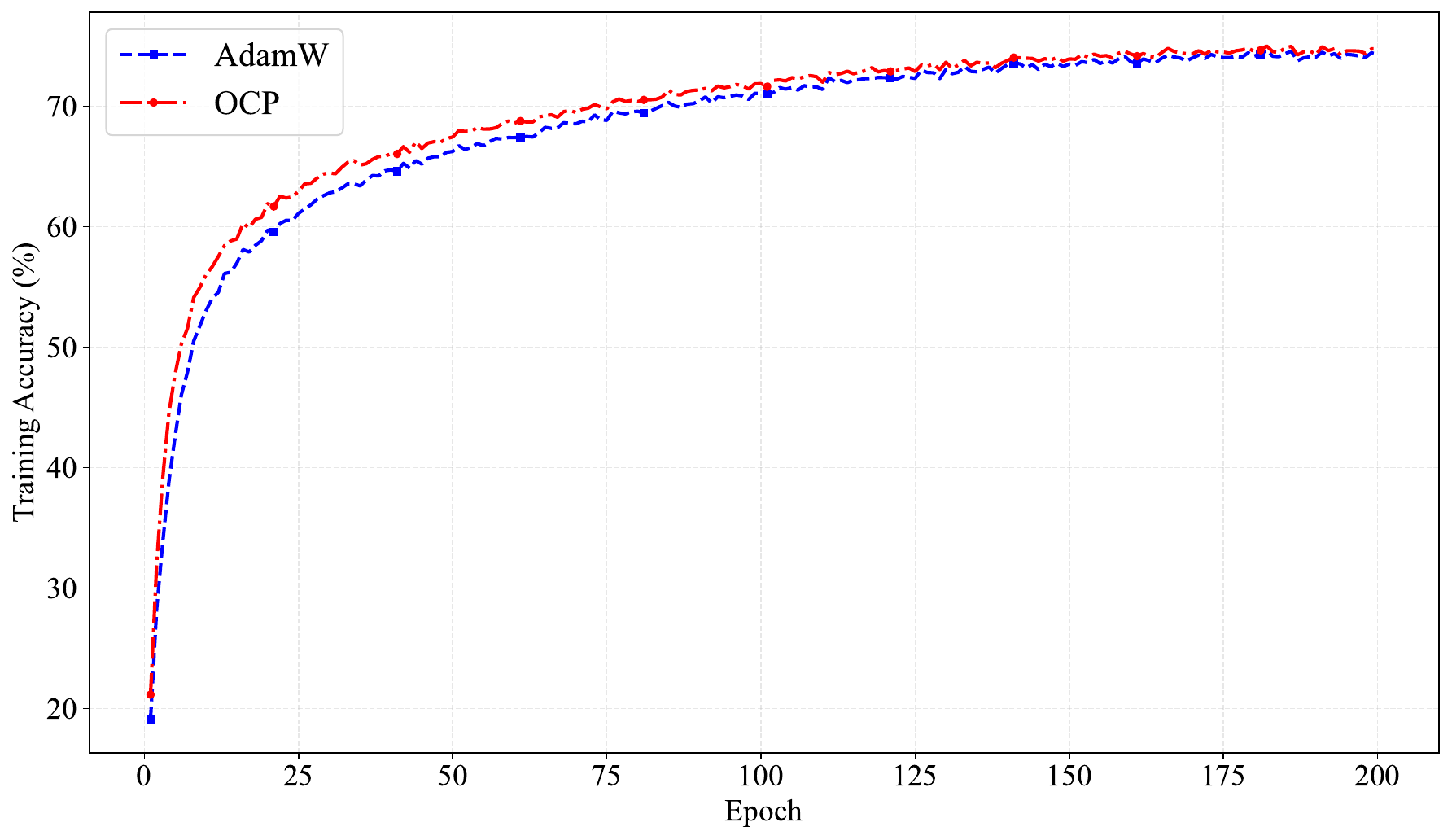}}
	\hspace{0.01\textwidth}
	\subfloat[ViT, CIFAR-100]{
		\includegraphics[width=0.23\textwidth]{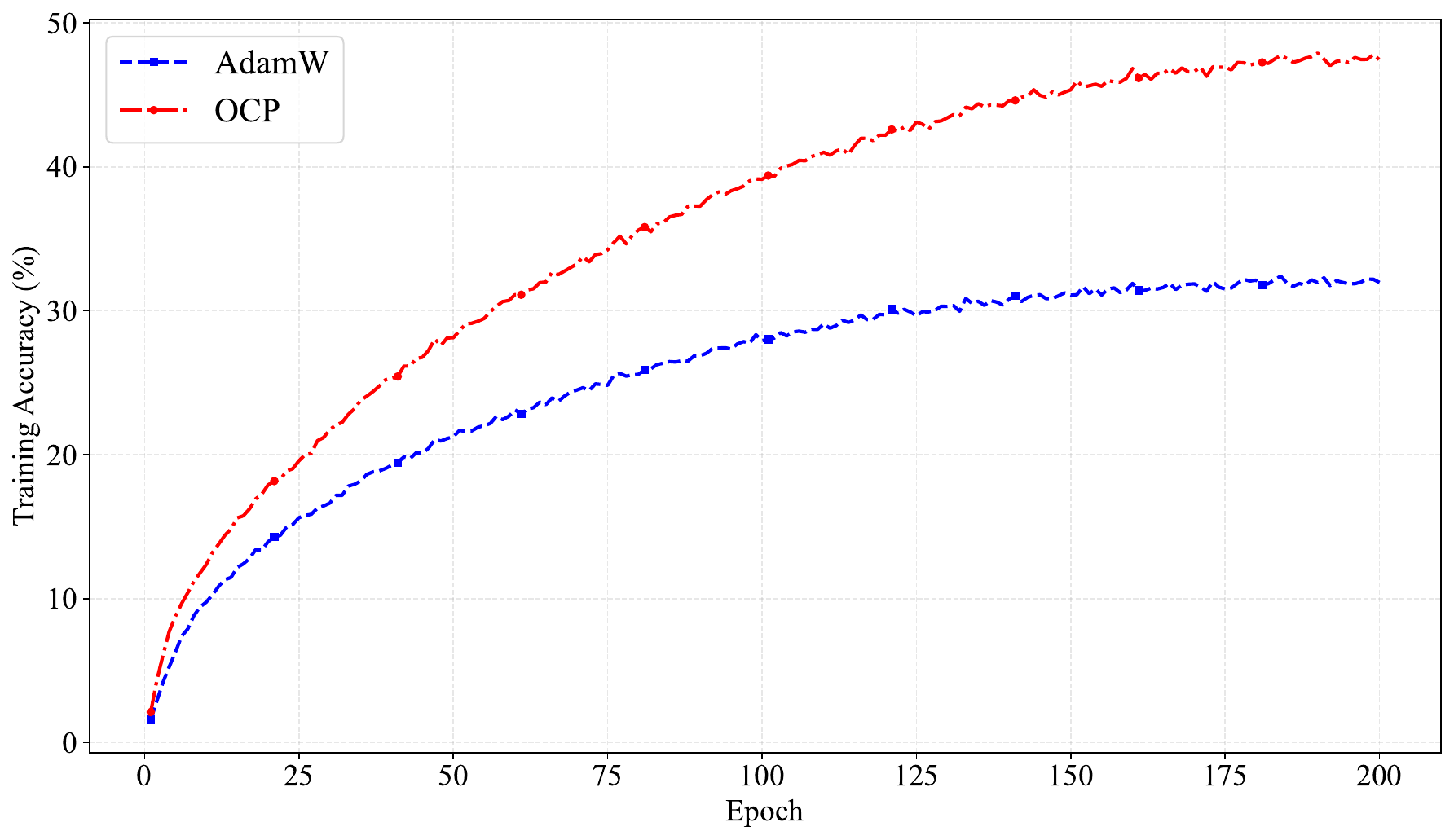}}
	\hspace{0.01\textwidth}
	\subfloat[ResNet-34, CIFAR-100]{
		\includegraphics[width=0.23\textwidth]{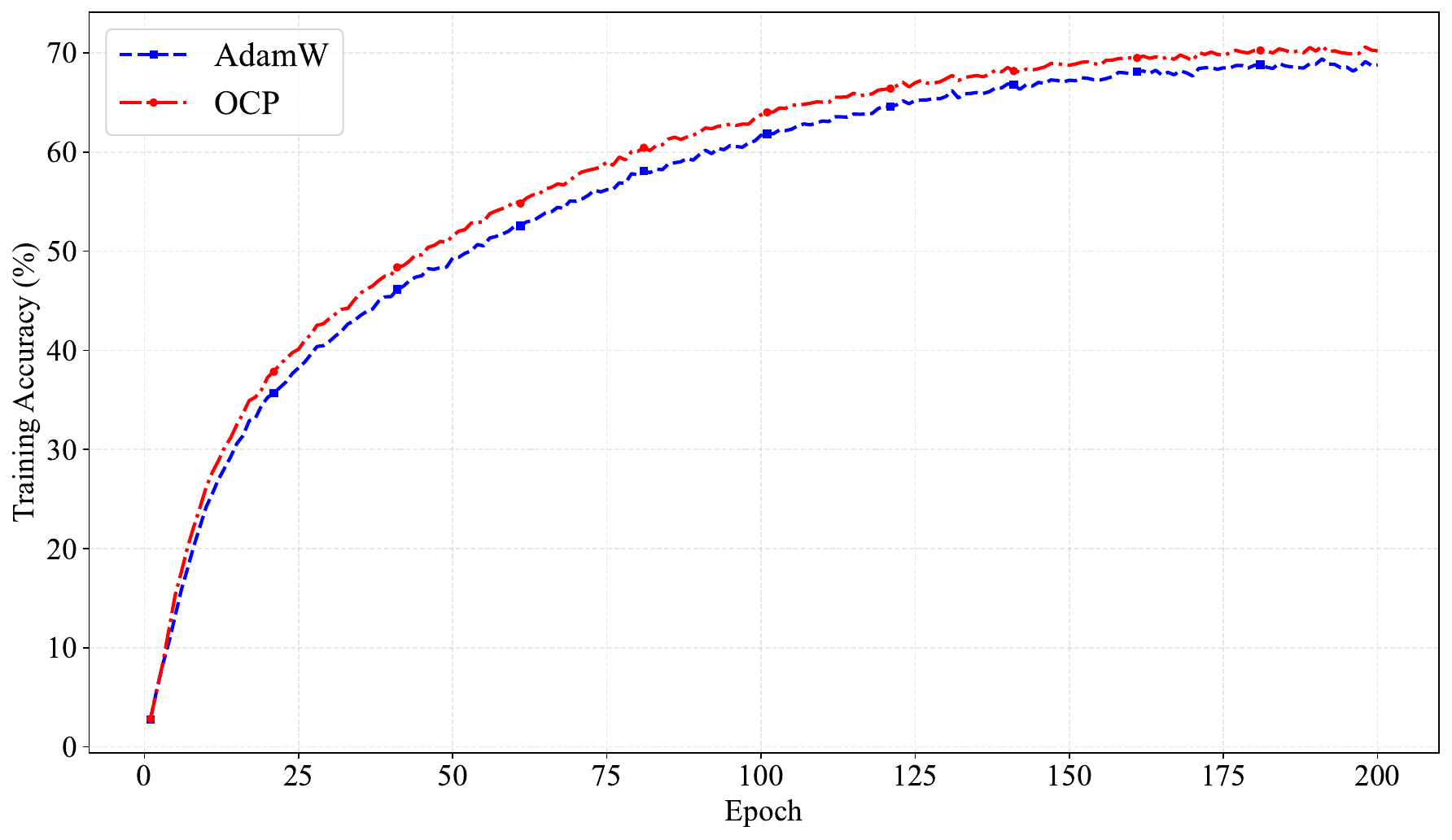}}
	
	\caption{Comparison of train accuracy between two algorithms.
		(a) ViT, CIFAR-10, (b) ResNet-34, CIFAR-10, (c) ViT, CIFAR-100, 
		(d) ResNet-34, CIFAR-100.}
		\label{fig:train_acc}
\end{figure*}

\begin{figure*}[htbp]
	\centering
	\subfloat[ViT, CIFAR-10]{
		\includegraphics[width=0.23\textwidth]{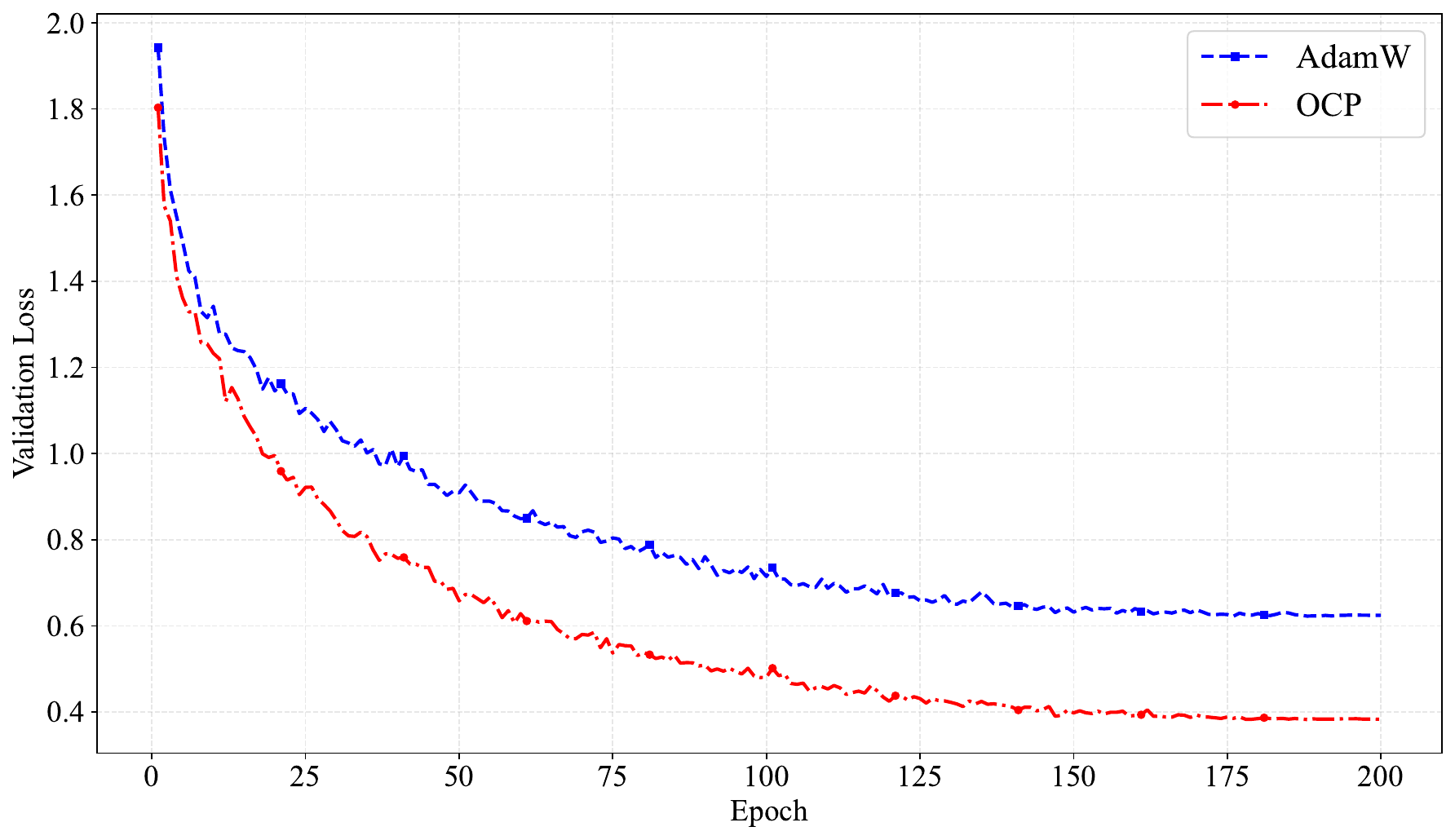}}
	\hspace{0.01\textwidth}
	\subfloat[ResNet-34, CIFAR-10]{
		\includegraphics[width=0.23\textwidth]{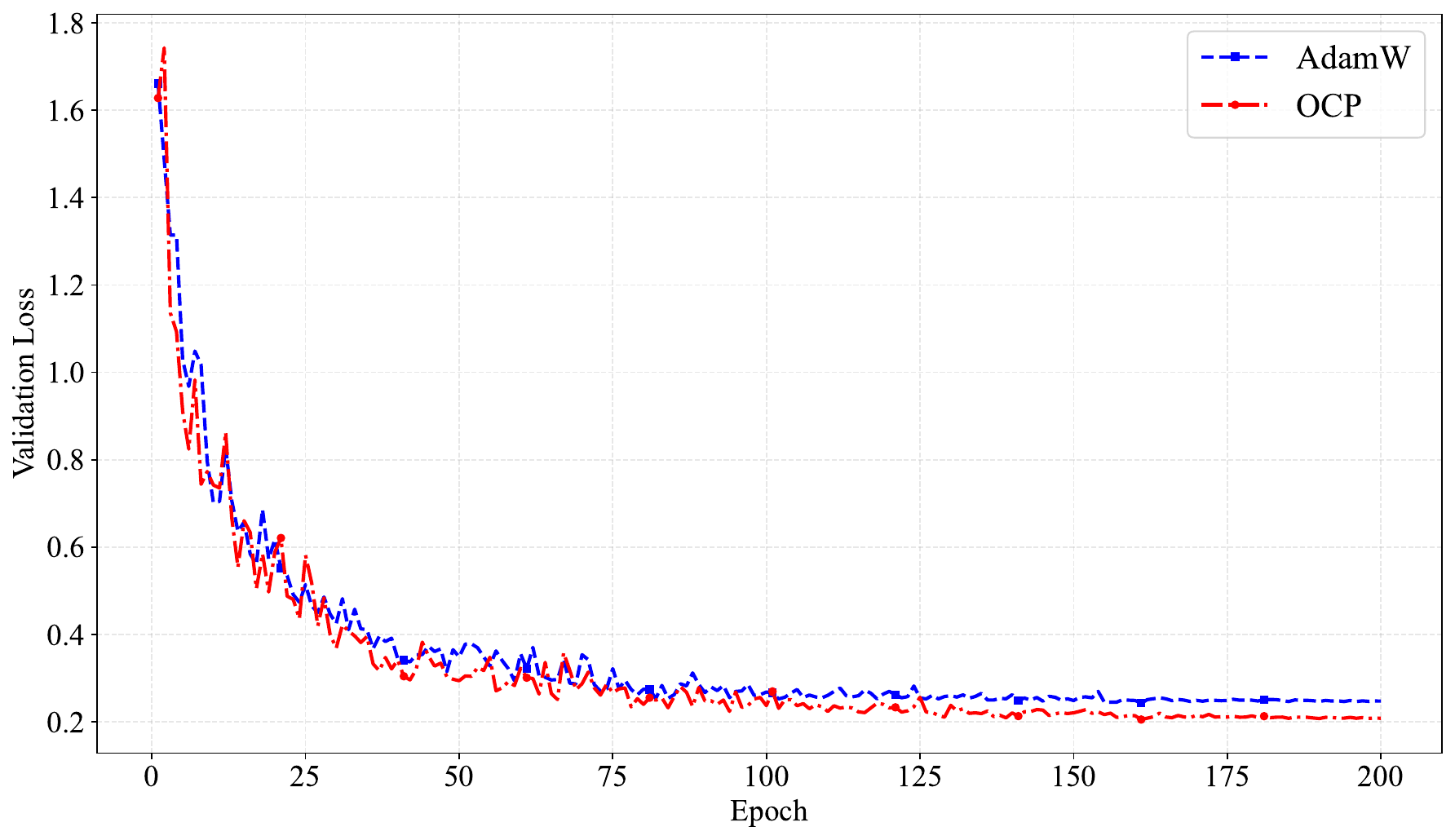}}
	\hspace{0.01\textwidth}
	\subfloat[ViT, CIFAR-100]{
		\includegraphics[width=0.23\textwidth]{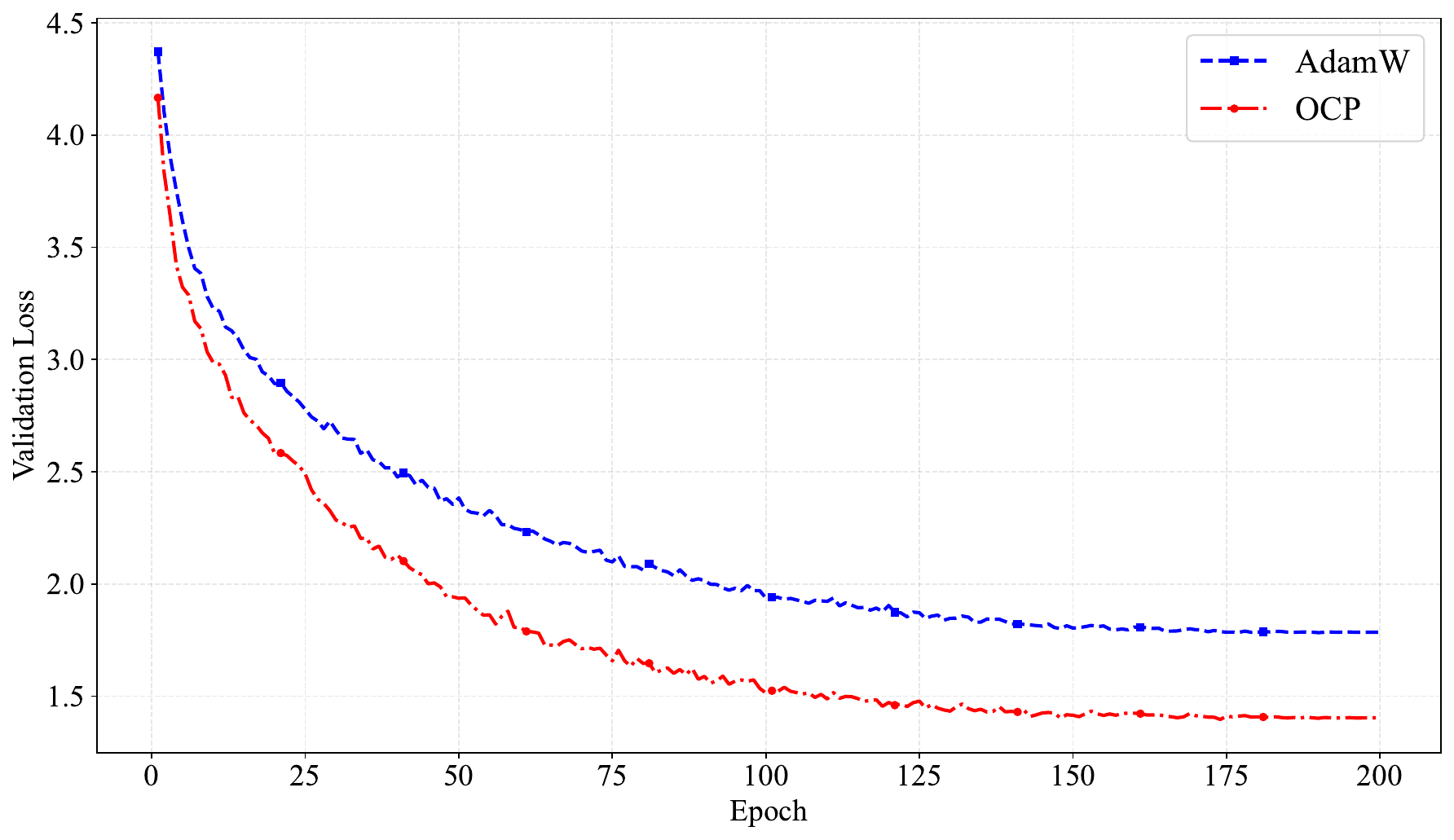}}
	\hspace{0.01\textwidth}
	\subfloat[ResNet-34, CIFAR-100]{
		\includegraphics[width=0.23\textwidth]{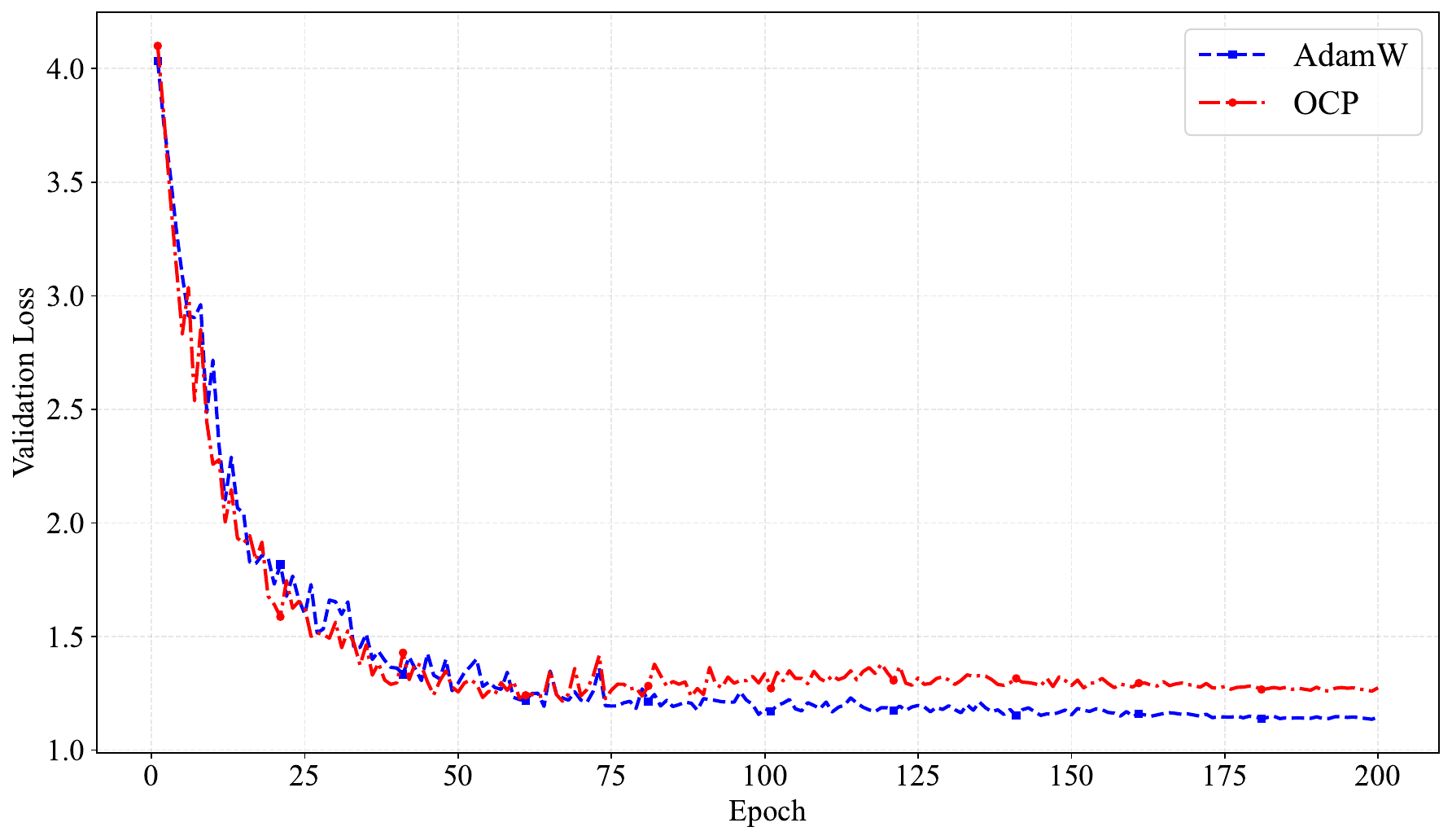}}
	
	\caption{Comparison of validation loss between two algorithms. 
		(a) ViT, CIFAR-10, (b) ResNet-34, CIFAR-10, (c) ViT, CIFAR-100, 
		(d) ResNet-34, CIFAR-100.}
		\label{fig:val_loss}
\end{figure*}

\begin{figure*}[htbp]
	\centering
	\subfloat[ViT, CIFAR-10]{
		\includegraphics[width=0.23\textwidth]{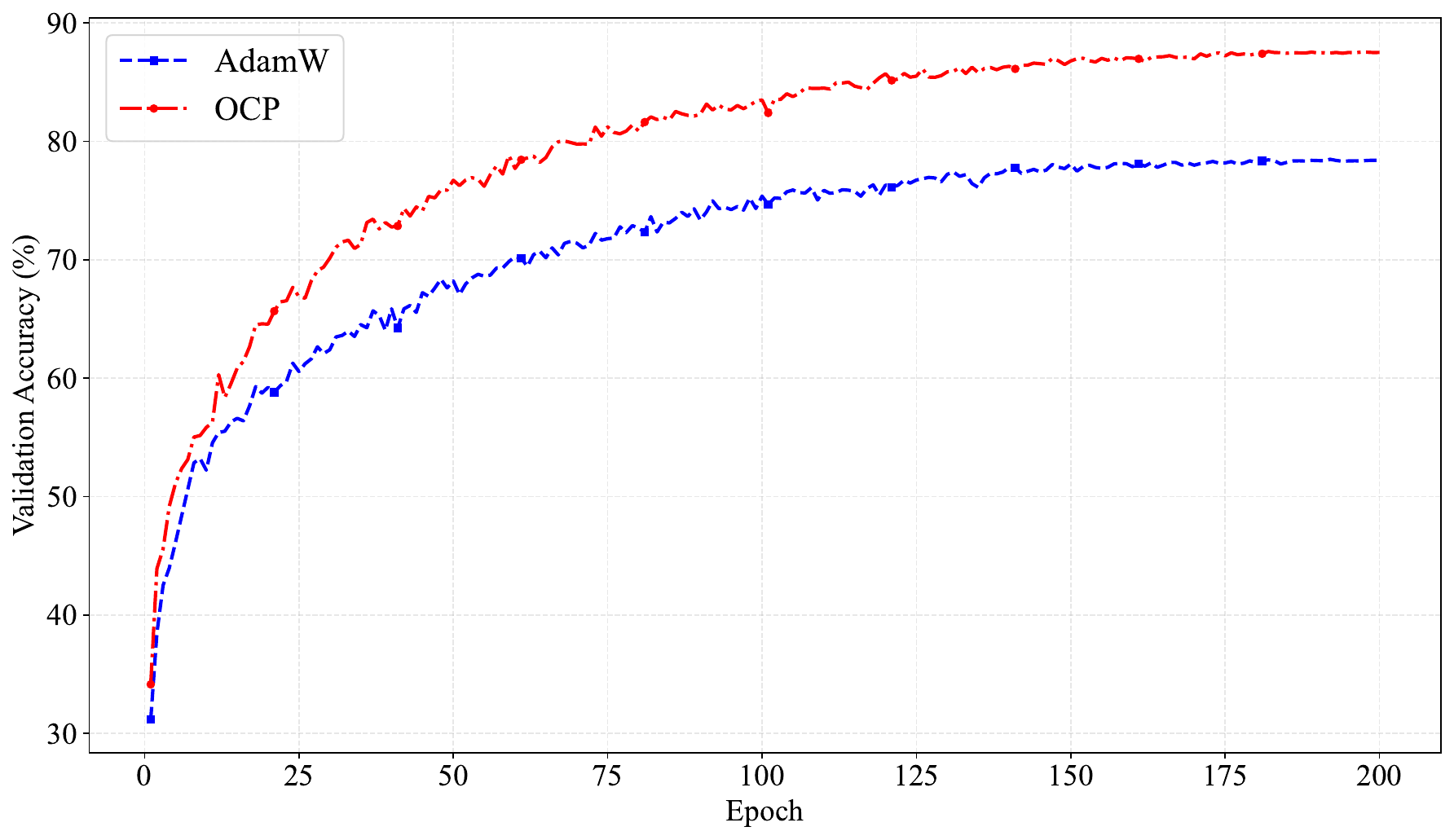}}
	\hspace{0.01\textwidth}
	\subfloat[ResNet-34, CIFAR-10]{
		\includegraphics[width=0.23\textwidth]{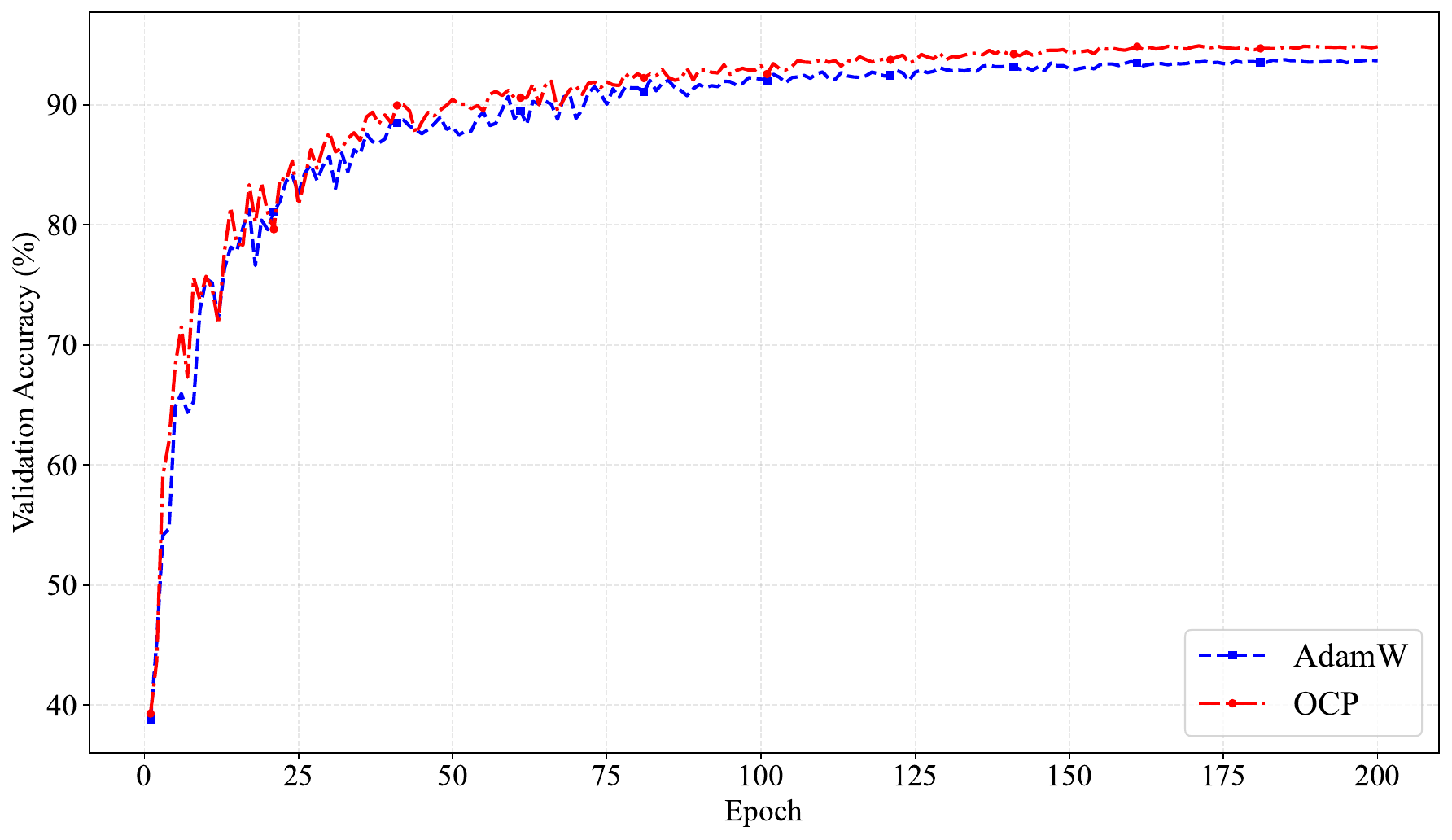}}
	\hspace{0.01\textwidth}
	\subfloat[ViT, CIFAR-100]{
		\includegraphics[width=0.23\textwidth]{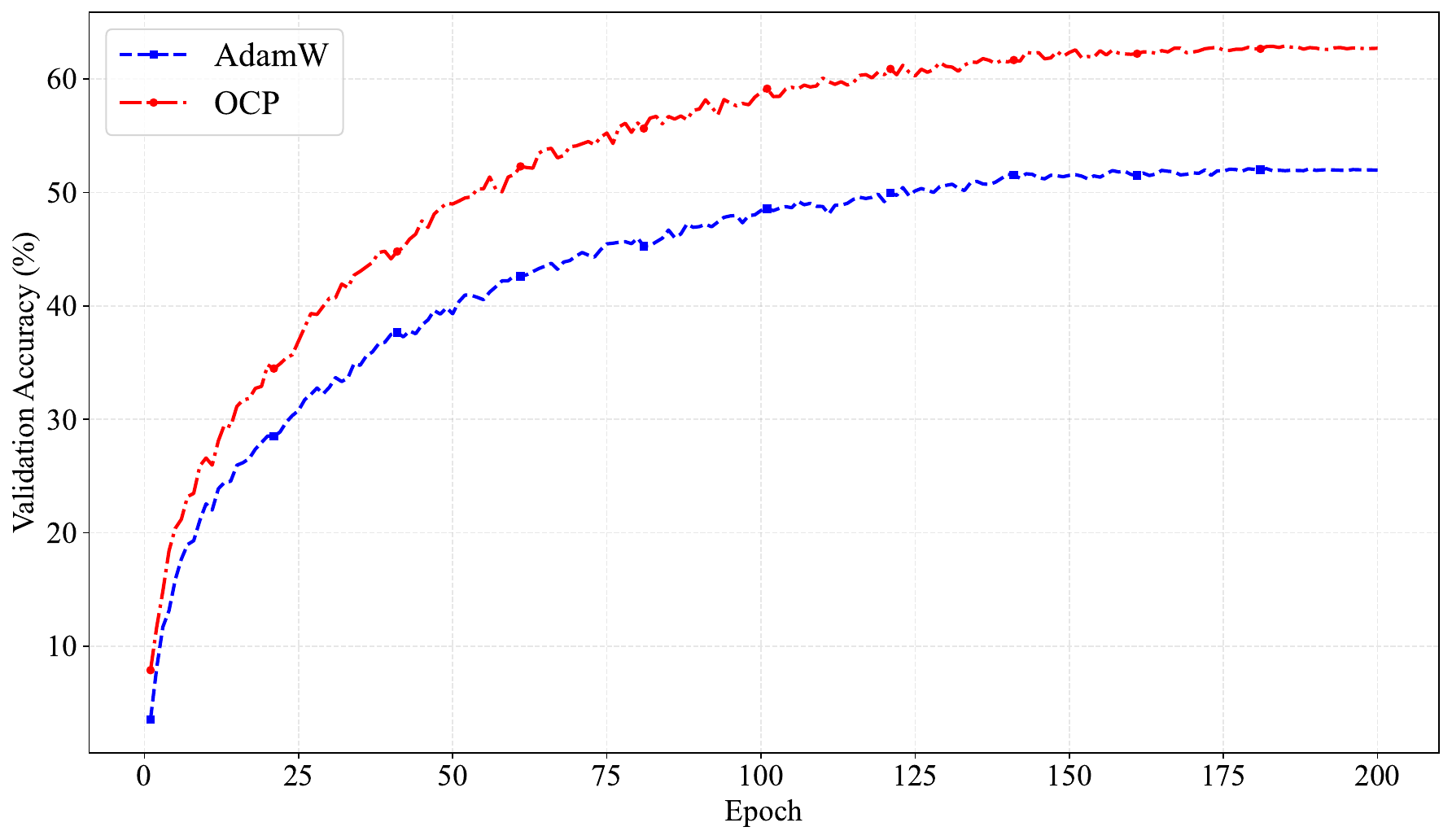}}
	\hspace{0.01\textwidth}
	\subfloat[ResNet-34, CIFAR-100]{
		\includegraphics[width=0.23\textwidth]{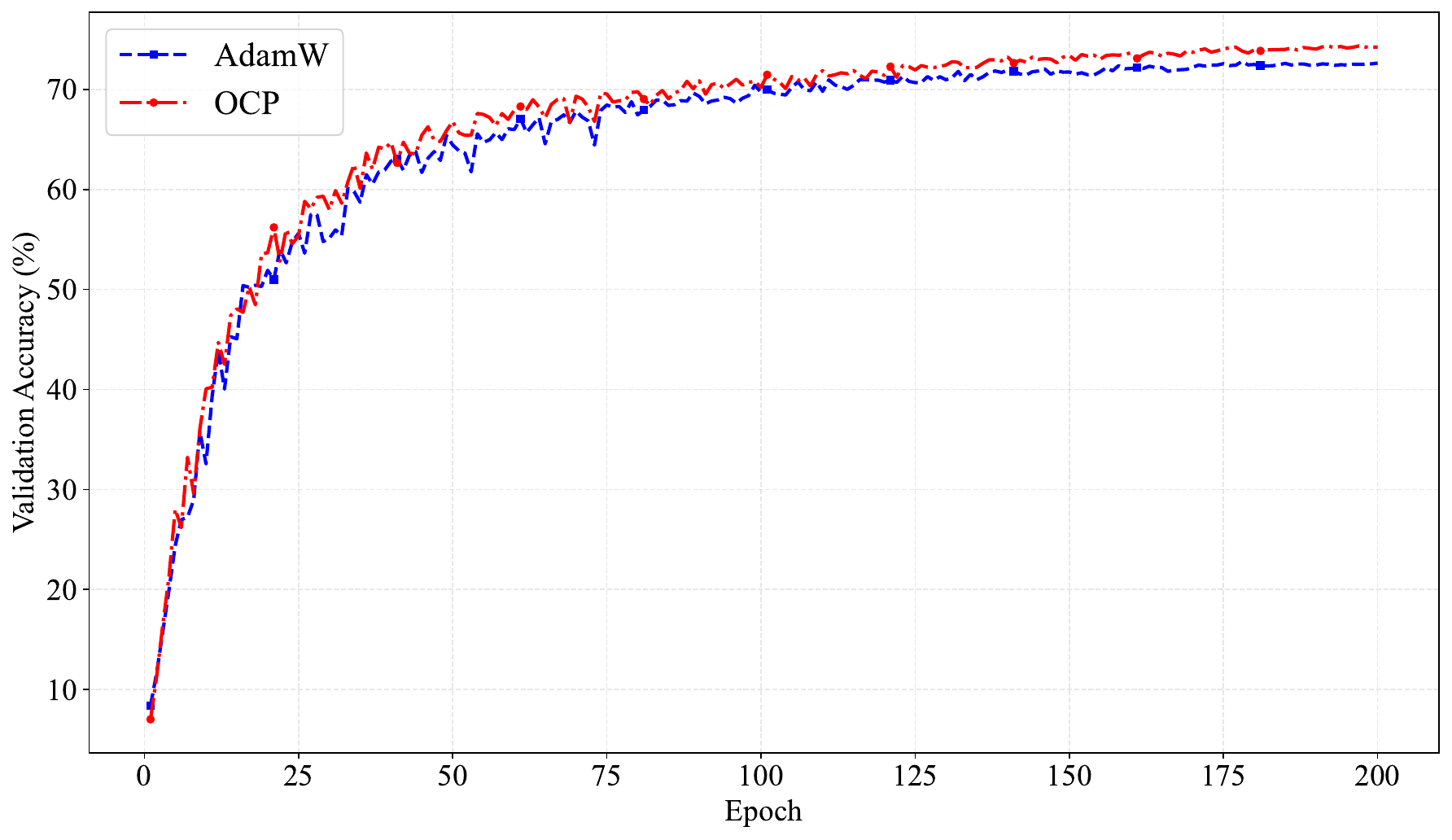}}
	
	\caption{Comparison of validation accuracy between two algorithms. 
		(a) ViT, CIFAR-10, (b) ResNet-34, CIFAR-10, (c) ViT, CIFAR-100, 
		(d) ResNet-34, CIFAR-100.}
		\label{fig:val_acc}
\end{figure*}

We validated the OCP-GN algorithm on image classification tasks and compared it with AdamW across different architectures, including ResNet-34 and Vision Transformer (ViT), as well as datasets of varying scales, including CIFAR-10 and CIFAR-100. As shown in \cref{fig:train_loss,fig:train_acc,fig:val_loss,fig:val_acc} and Table \ref{tab:classification}, OCP-GN outperforms AdamW in both convergence rate and test performance.
\begin{table}[htbp]
	\centering
	\caption{Performance metrics of algorithms on the classification tasks}
	\label{tab:classification}
	{\fontsize{8}{10}\selectfont 
		\setlength{\tabcolsep}{6pt}
		\begin{tabular}{ c | c c | c c }
			\toprule
			Dataset & \multicolumn{2}{c|}{CIFAR-10} & \multicolumn{2}{c}{CIFAR-100} \\
			Model & ViT & ResNet-34 & ViT & ResNet-34 \\
			\midrule
			OCP-GN & \textbf{87.50\%} & \textbf{94.84\%}  & \textbf{62.70\%} & \textbf{74.22\%} \\
			
			AdamW & 78.39\% & 93.67\%  & 51.96\% & 72.64\% \\
			\bottomrule
		\end{tabular}
	}
\end{table}

\section{Conclusion}
We propose OCP-GN from OCP method. The optimizer inherits the update rule of OCP method, while using a GNB estimator to simplify the computation of the Hessian matrix and incorporating multiple technical improvements for stable updates. In image classification tasks, experiments on various model architectures and datasets validated the significant advantages of the proposed optimizer in terms of convergence performance and generalization capability.


\begin{thebibliography}{00}
	\bibitem{posenet} Kendall A, Grimes M, Cipolla R. Posenet: A convolutional network for real-time 6-dof camera relocalization[C]//Proceedings of the IEEE international conference on computer vision. 2015: 2938-2946.
	
	\bibitem{mapnet} Brahmbhatt S, Gu J, Kim K, et al. Geometry-aware learning of maps for camera localization[C]//Proceedings of the IEEE conference on computer vision and pattern recognition. 2018: 2616-2625.
	
	\bibitem{loss fuction} Kendall A, Cipolla R. Geometric loss functions for camera pose regression with deep learning[C]//Proceedings of the IEEE conference on computer vision and pattern recognition. 2017: 5974-5983.
	
	\bibitem{ocp1} Zhang H, Wang H, Xu Y, et al. Optimization methods rooted in optimal control[J]. Science China Information Sciences, 2024, 67(12): 222208.
	
	\bibitem{ocp2} Wang H, Xu Y, Guo Z, et al. Optimization Algorithms with Superlinear Convergence Rate[J]. IEEE Transactions on Automatic Control, 2025.
	
	\bibitem{ocp-ls} Zhong J, Guo Z, Wang H, et al. An Efficient Algorithm for Learning-Based Visual Localization[J]. arXiv preprint arXiv:2511.04232, 2025.
	
	\bibitem{sophia} Liu H, Li Z, Hall D L W, et al. Sophia: A Scalable Stochastic Second-order Optimizer for Language Model Pre-training[C]//The Twelfth International Conference on Learning Representations.
\end{thebibliography}
\end{document}